\documentclass{article}
\usepackage{ijcai22}

\usepackage{times}
\usepackage{soul}
\usepackage{url}
\usepackage[hidelinks]{hyperref}
\usepackage[utf8]{inputenc}
\usepackage[small]{caption}
\usepackage{graphicx}
\usepackage{amsmath}
\usepackage{amsthm}
\usepackage{amssymb}
\usepackage{booktabs}
\usepackage{algorithm}
\usepackage{algorithmic}
\usepackage{color}
\usepackage{multirow}
\usepackage{subcaption}
\title{$I^2$R-Net: Intra- and Inter-Human Relation Network \\for Multi-Person Pose Estimation} 

\author{
Yiwei Ding$^1$\thanks{Equal Contribution.}
\and
Wenjin Deng$^{1*}$\and
Yinglin Zheng$^{1}$\and
Pengfei Liu$^1$\and\\
Meihong Wang$^1$\and
Xuan Cheng$^1$\and
Jianmin Bao$^2$\and
Dong Chen$^2$\and
Ming Zeng$^{1}$\thanks{Corresponding Author.}
\affiliations
$^1$School of Informatics, Xiamen University\\
$^2$Microsoft Research Asia \\
\emails
\{dingyiwei,dengwenjin,zhengyinglin,pengfei\}@stu.xmu.edu.cn,
\{wangmh,chengxuan,zengming\}@xmu.edu.cn,\{jianbao, doch\}@microsoft.com
}

\begin{document}

\maketitle

\begin{abstract}

In this paper, we present the Intra- and Inter-Human Relation Networks~(\textbf{$I^2$R-Net}) for Multi-Person Pose Estimation. It involves two basic modules. First, the Intra-Human Relation Module operates on a single person and aims to capture Intra-Human dependencies. Second, the Inter-Human Relation Module considers the relation between multiple instances and focuses on capturing Inter-Human interactions. The Inter-Human Relation Module can be designed very lightweight by reducing the resolution of feature map, yet learn useful relation information to significantly boost the performance of the Intra-Human Relation Module.
Even \emph{without bells and whistles}, our method can compete or outperform current competition winners. We conduct extensive experiments on COCO, CrowdPose, and OCHuman datasets. The results demonstrate that the proposed model surpasses all the state-of-the-art methods. Concretely, the proposed method achieves \textbf{77.4\%} AP on CrowPose dataset and \textbf{67.8\%} AP on OCHuman dataset respectively, outperforming existing methods by a large margin. Additionally, the ablation study and visualization analysis also prove the effectiveness of our model.
  
\end{abstract}

\vspace{-3mm}
\section{Introduction}
2D Multi-person Pose Estimation (MPPE) aims to detect and localize the human keypoints for all persons appearing in a given image. Since human pose provides abundant structural and motion information, MPPE has attracted attentions in wide applications, such as human activity understanding, human-object interaction, avatar animation, etc.

\begin{figure}[t]
 \centering
    \includegraphics[width=0.48\textwidth]{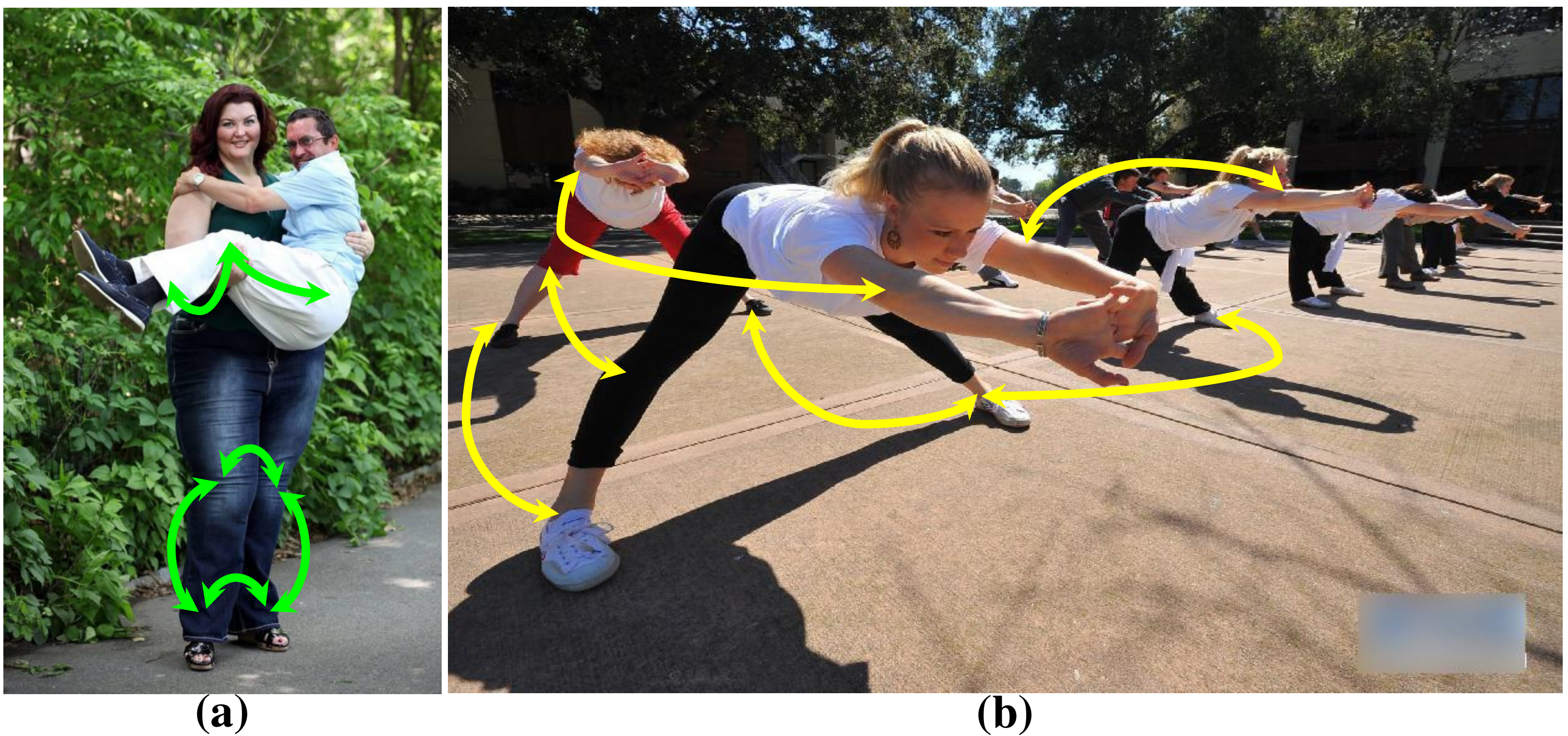}
    \vspace{-4mm}
  \caption{\textbf{Illustrations of Intra- and Inter-Human Relations.} 
  (a) the intra-human relation is marked as the green arrow~(left-right pairs and two keypoints of the same limb), while the inter-human relation is denoted as the yellow arrow~(the same type of keypoints in different individuals).} 
  \label{head}
  \vspace{-3mm}
\end{figure}

Current MPPE methods can be divided into two fashions, \emph{i.e.}, top-down and bottom-up. Top-down methods \cite{he2017mask,xiao2018simple,fang2017rmpe,sun2019deep,chen2018cascaded} detect bounding box for each human, then estimate the pose of each human separately. Bottom-up methods \cite{cao2019openpose,newell2016associative,cheng2020higherhrnet,papandreou2018personlab,jin2020differentiable} localize all joints in the image, and then group them into individuals.
Despite having achieved great performance in MPPE, the above methods produce inferior results in crowded scenes, due to overlapping, self-occlusion, and a large variety of appearances and poses. Top-down methods treat each person separately, neglecting the informative cues of mutual interaction among persons. On the other hand, bottom-up methods jointly detect joints of all persons in an image, potentially correlating joints across persons, but the relationships between persons are vague and agnostic. Recently, several pioneers address the above challenges via decoupling poses \cite{li2019crowdpose,qiu2020peeking} or fusing multiple possible poses \cite{wang2021robust} at a single person. However, these methods still do not fully take the multiple-human correlations into account, leaving these important interactions unexplored.

This work investigates how to leverage the correlations of intra-human or cross-human, \emph{e.g.}, persons with similar poses or persons who closely interact with each other, to improve the accuracy of multi-person pose estimation. As shown in Figure~\ref{head}(a), two people are hugging each other. Their poses show strong intra-human information~(\emph{e.g.}, left-right pairs and two keypoints of the same limb) and obvious inter-human relations. In  Figure~\ref{head}(b), these persons are dancing together, indicating that their poses are similar, also providing inter-human cues for more accurate pose estimation, especially in the occluded regions or low-resolution regions.


Based on above observations, we propose a novel two-stage Intra- and Inter-Human Relation Networks~(\textbf{$I^2$R-Net}) for MPPE. The first stage targets at learning the dependencies among body parts of a single person. It operates at high resolution and generates the locations of small parts, such as the eyes, wrists, and ankles. The second stage operates multi-person collaboratively at low resolution. It aims to capture interactions between instances, like touching hands, joining arms, and crossing legs. It's worth noting that each instance is represented by a low-resolution feature map rather than merely a vector. By doing this, we can obtain variant responses from regions, since the cross-instance interaction depends on spatial semantic information. Meanwhile, the low-resolution feature map significantly reduces the computation cost of the second stage but is sufficient to capture useful correlation information. 
We call the first \textbf{Intra-Human Relation Module}, and the second \textbf{Inter-Human Relation Module}. The two stages are sequentially stacked with skip-connection.


This two-stage framework is structure flexible and function specialized for MPPE. The Intra-Human Relation Module can be an arbitrary single-person pose estimation method, aiming to explore part-level patterns for keypoint detection. Moreover, the Inter-Human Relation Module can be implemented by any non-local model, to pay more attention to correlations in semantic features. The two modules are able to work collaboratively to infer multi-person poses, because the first module offers high-quality pose information to the next module, while the second module aids the first to eliminate ambiguities in occluded or less discriminative regions.

We evaluate our method on common benchmarks for crowded scenes including CrowdPose and OCHuman. Experiments have proved that our method surpasses all state-of-the-art approaches by large margins. We achieve 77.4\% AP on CrowdPose, outperforming HRFormer-B by 5.0\%. We achieve 67.8\% AP on OCHuman dataset, outperforming TransPose-H by 5.5\%. 
In addition, we bring improvements on COCO dataset, which means our method can generally work in non-crowded scenes. 

In summary, the contributions of our work are as follows:

\begin{itemize}

\item We propose a two-stage MPPE framework that not only relates each body part in a single person, but also builds connections among multi-person in the images. The proposed method bridges the gap between pose estimation and human-human interactions.

\item The framework is designed to be flexible. The module coping with intra-human relations can be arbitrary pose estimation method. While the module that models inter-human relations is very lightweight.

\item Extensive experiments show that, without bells and whistles, our method significantly surpasses state-of-the-art methods on challenging datasets like CrowdPose, OCHuman, and COCO.
\end{itemize}

\section{Related Work}
\subsection{Multi-Person Pose Estimation}

\paragraph{Top-down Framework.}
Thanks to the great success of object detection,
it is intuitive to estimate the pose of each person with the given human bounding box. Thus, lots of work under top-down framework focus on developing single-person estimation without optimizing the off-the-shelf detection models.
Reviewing the methods in recent years, the major development idea is to improve the feature representation in spatial. \cite{he2017mask} directly adds a keypoint detection branch on a CNN-based feature extractor. \cite{newell2016stacked} follows multi-stage fashion to stack modules for dense prediction. Furthermore, applied with intermediate supervision at each stage, the model is driven to perform estimation from coarse to fine. CPN~\cite{chen2018cascaded} employs pyramid structure in spatial to handle parts on different scales. Instead of deepening network with sequential modules, HRNet~\cite{sun2019deep} maintains multi-resolution in parallel leading to rich high-resolution representation.\\

\noindent\paragraph{Bottom-up Framework.}
Bottom-up methods first predict all identity-free body keypoints from an uncropped image, then group these body keypoints into different individuals. Most of them adopt heatmap for keypoint detection, and focus on how to identify and group keypoints to the corresponding person. OpenPose \cite{cao2019openpose} proposed the part affinity field to reveal the relationship between two keypoints in the same limb. During the grouping stage, after calculating the line integral score between two candidate keypoints, the pair with the highest score is associated. This approach is denoted as PAF family. Another popular fashion is Associative Embedding, called AE\cite{newell2016associative,cheng2020higherhrnet}. They learn the tag embeddings for each person and group keypoints into individuals by clustering the tag. 

Top-down methods estimate poses one by one, without taking the relationship between instances into account. While bottom-up methods detect all poses together, the relationship cues are vague. In contrast, our work pays attention to capturing the hierarchical interactions in the multi-person scenario to promote pose estimation performance. Thus, we provide a novel two-stage framework to focus on modeling part-level and instance-level relationships for MPPE.\\

\noindent\paragraph{Human Pose Estimation in Crowd.}
The performance of previous work decrease as the number of human increases.  Even though top-down methods achieve higher performance than bottom-up methods, they still fail in crowded scenes due to occlusions. Thus, CrowdPose dataset \cite{li2019crowdpose} and OCHuman dataset \cite{zhang2019pose2seg} have been proposed to encourage researchers to study this challenging problem. 
Existing methods \cite{li2019crowdpose,qiu2020peeking} attempt to decompose coupled poses in crowd. 

In this paper, however, the proposed Inter-Human Relation Module is designed to conquer this dilemma by leveraging the relationships between instances.

\subsection{Vision Transformer}
Transformer is an attention-based model, and has been studied in Neural Language Processing in recent years, like BERT~\cite{devlin2018bert} and GPT~\cite{radford2019language}. With the effectiveness of Transformer, recently, an increasing number of works~\cite{parmar2018image,bello2019attention,touvron2021training} use Vision Transformer for computer vision tasks. 
Several previous works apply Transformer to 2D pose estimation. For example, Transpose ~\cite{yang2021transpose} proposes a pose estimator with attention layers to capture and reveal the spatial dependencies between 2D heatmaps. HRFormer \cite{YuanFHLZCW21} learns high-resolution representations via self-attention schema. 
However, none of them take human relations into account.

Although recent work \cite{wang2021multi,Mihai2021rempis} utilize Transformer to calculate relations, yet they aim to solve 3d motion prediction or reconstruction where the relationships between different persons are intuitive. In contrast, our work is the first to incorporate multi-person correlations in single image 2D Pose Estimation, where the relationship is hard to build and we propose a novel module to solve this. Moreover, instead of using a well-designed framework for only building inter-human relationships, we propose a flexible, lightweight, and general module for this purpose that can be easily integrated into current or future single-person pose estimation frameworks.

\begin{figure}[t]
 \centering
    \includegraphics[width=1.0\linewidth]{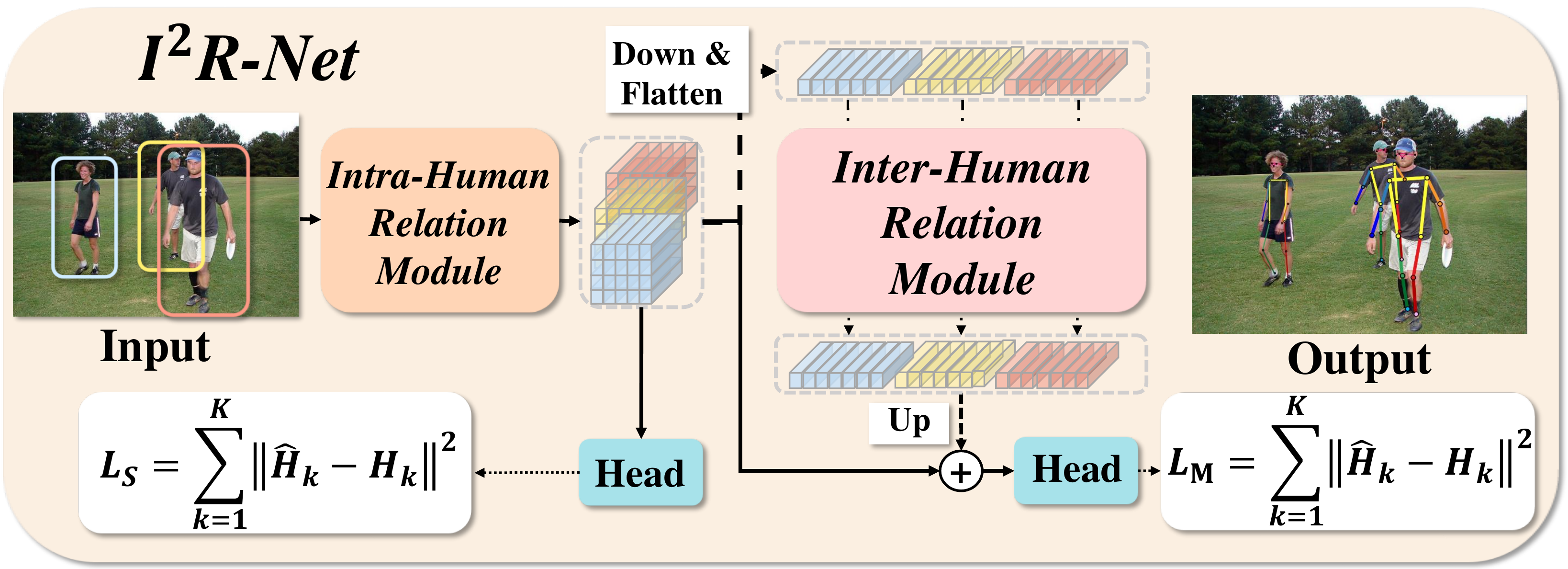}
  \caption{\textbf{$I^2$R-Net Framework.} The framework includes an Intra-Human Relation Module and an Inter-Human Relation Module. The Intra-Human Relation Module can apply current single-person pose estimation methods flexibly. The Inter-Human Relation Module is lightweight by downsampling the feature from Intra-Human Relation Modules.} 
  \label{framework}
\end{figure}


\section{Approach}

\subsection{Overview of $I^2$R-Net}

As shown in Figure~\ref{framework}, the framework takes the detected and cropped human image patches in the image as input. In the first stage, Intra-Human Relation Network~(Section~\ref{IntraModule}) extracts the feature of each human patch separately and preserves relatively high-resolution feature representation. In the second stage, Inter-Human Relation Module~(Section~\ref{InterModule}) exchanges information of the per-person features extracted by the first stage, and the per-person features should be downsampled beforehand to reduce computation cost. Finally, the two stages are stacked with a skip-connection to fuse the high-resolution intra-human information and the low-resolution inter-human information.

\subsection{Intra-Human Relation Module}\label{IntraModule}
As the goal of Intra-Human Relation Module is to learn semantic representations of the human patches and capture the dependencies between body parts, it operates at patches of cropped human images. It can be an arbitrary single-person human pose estimation method. One key factor is the input and output resolutions, which is a trade-off between performance and computational consumption. Following the recent works~\cite{yang2021transpose,YuanFHLZCW21}, we adopt two settings, \emph{i.e.}, $256\times192$ of input and $64\times48$ of output as default, and $384\times288$ of input and $96\times72$ of output for comparison.

\subsection{Inter-Human Relation Module}\label{InterModule}

\begin{figure}[t]
 \centering
    \includegraphics[width=0.48\textwidth]{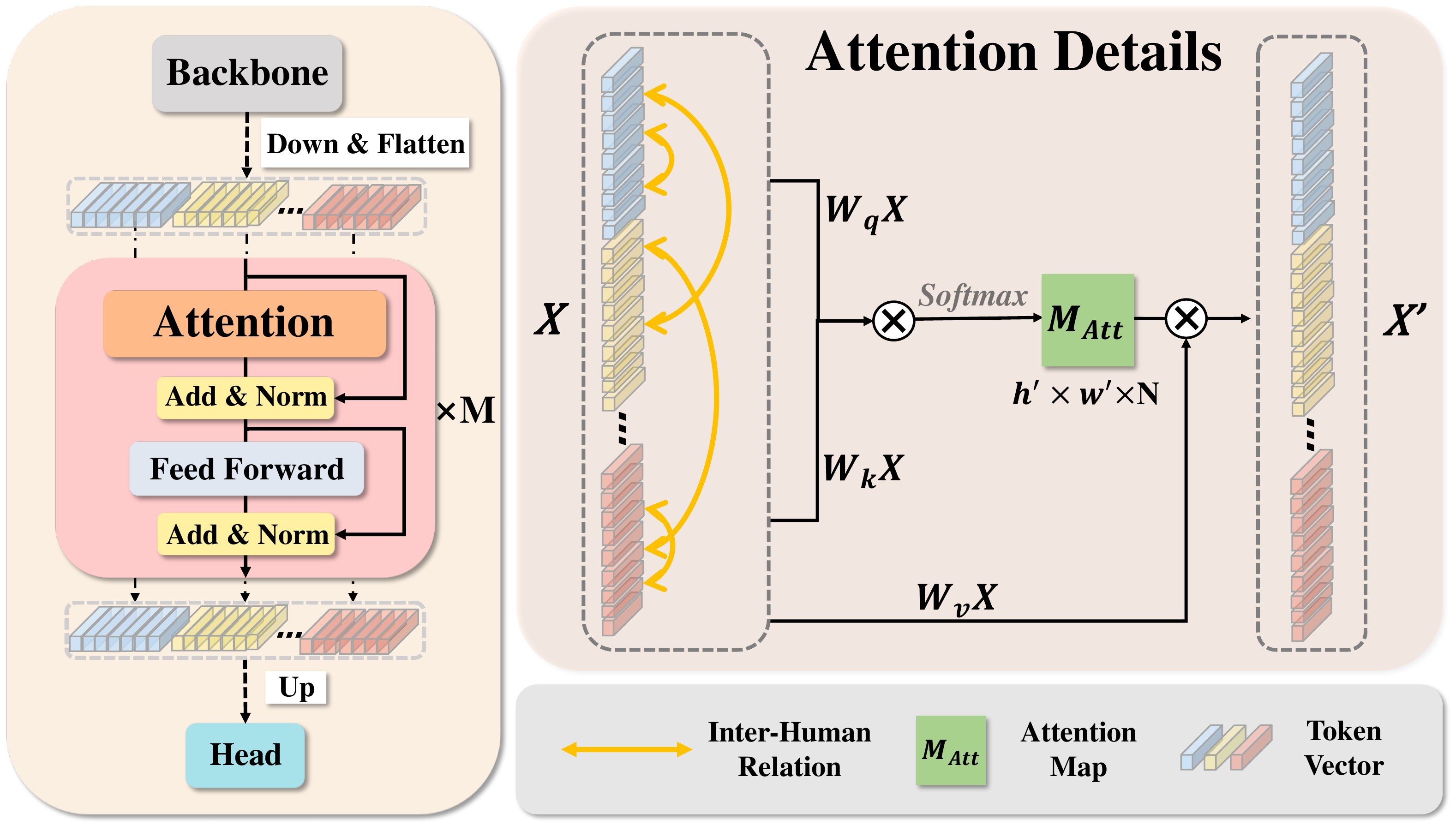}
  \caption{\textbf{A Vanilla Inter-Human Network.} The vanilla network uses Inter-Human Relation Module only with a conventional CNN-based model to extract features of each person cropped from the input image. The extracted features are  downsampled, flattened, and concatenated before being passed into Inter-Human Relation Module. The output features of the Inter-Human Relation Module are upsampled to higher resolution, and then fed into a keypoint prediction head.}
  \label{interformer}
\end{figure}



For the task of Multi-Human Pose Estimation, we devise an Inter-Human Relation Module to collaboratively infer the pose of each person via explicitly modeling cross-human correlation. 

The Inter-Human Relation Module is lightweight yet effective. We illustrate a vanilla network example in Figure~\ref{interformer}. Given an image $I$ with $N$ persons, the Intra-Human Relation Module (\emph{e.g.}, HRNet, TransPose) locates all persons and extracts their features $\mathbf{P_1}, \mathbf{P_2},..., \mathbf{P_N}$ from raw patches. These features contain pose information of each person. The Inter-Human Relation Module takes these features as input and exchanges pose information among each instance. To fully capture the correlations between different people, we adopt the recently popular Transformer blocks.
It is worth noting that, as shown in Figure~\ref{interformer}, the Inter-Human relation module not only exchanges information across instances, but also correlates information among parts in each instance. This ensures that even using only an Inter-Human Relation Module~(without the Intra- stage, and the input per-person features are extracted by a backbone), it can still achieve state-of-the-art performance in our experiments. 


Concretely, suppose the output feature maps of first stage is $\mathbf{P_i} \in \mathbb{R}^{h\times w\times d}, i \in [1, N]$, where $h \times w$ is the feature map resolution, $d$ is the channel dimension, $N$ is the number of persons. We use the max pooling operator with kernel size and stride set to $R$ to operate on $\mathbf{P_i}$ and get resulted downsampled feature map $\mathbf{P_i}' \in \mathbb{R}^{h/R\times w/R\times d}$. Then we flatten all the features in $\mathbf{P_i}'$ and concatenate them to get a sequence with a length of $L = h/R\times w/R \times N$: 


\begin{equation}
\mathbf{P_i}' = \text{Downsample}(\mathbf{P_i})
 \end{equation}
\begin{equation}
    \mathbf{X} = \text{Concat}[\text{Flatten}(\mathbf{P_i}')], i \in [1, N] 
\end{equation}
where $\textbf{X}\in\mathbb{R}^{L\times d}$. 

Subsequently, $\textbf{X}$ is fed into $M$~(set to 6 for vanilla network) stacked Transformer blocks, aiming to compute correlations between different instances. The formulation of attention operation is given as follows, 
    
\begin{equation}
    \text{Attention}(X) = \text{Softmax}\left[\frac{(\textbf{W}_q\textbf{X})(\textbf{W}_k\textbf{X})^T}{\sqrt{d}}\right]\textbf{W}_v\textbf{X}
\end{equation}
where $\textbf{W}_q,\textbf{W}_k,\textbf{W}_v \in \mathbb{R}^{d\times d}$ are learnable parameters.

Finally, we use a standard deconvolution layer to upsample the output embeddings  $\textbf{X}'$ to higher resolution and employ a keypoint head to predict $K$ types of keypoint heatmaps $\hat{\textbf{H}}$, \emph{i.e.}, $\hat{\textbf{H}} = \text{Head}(\text{Upsample}(\textbf{X}'))$.

Following the previous methods, we adopt the MSE loss to supervise predicted keypoints heatmaps $\hat{\textbf{H}}$ with the ground-truth $\textbf{H}$, \emph{i.e.}, $\sum_{k=1}^{K}{\left \| \hat{\textbf{H}}_k-\textbf{H}_k \right \|}^2$. In our framework, the Intra-Human loss $\mathcal{L_S}$ serves as an intermediate supervision. Thus, after summing up instance-level stage loss $\mathcal{L_M}$, the total loss is $\mathcal{L}= \mathcal{L_S}+\alpha \mathcal{L_M}$. In our experiments, we set $\alpha=1$.


Our proposed Inter-Human Relation Module is general and flexible, and it can be instantiated with different 
attention methods~(\emph{e.g.}, Non-local networks, Transformer Layers) and implementation specifics. 

\noindent\textbf{Input of Inter-Human Relation Module.}
To reduce the computation complexity, we downsample the output feature map of the Intra-Human Relation module with the ratio $R$ and restrict the number of persons to $N$. Suppose the input feature map size is $h \times w$, the downsampled feature map is $h'\times w'$, then we flatten and concatenate all these elements together. After that,  we perform self-attention calculations on all these feature elements~(shown as Figure~\ref{interformer}). The correlations between poses of different persons are captured. Since $N$ is an important hyper-parameter for the computation complexity. We set a max person number $N$ for each input image differently on different datasets. Specifically, we adopt $N=6$ for CrowdPose, $N=3$ for OCHuman and $N=4$ for COCO, respectively. Experimental results prove that our input setting of the Inter-Human Relation Module achieves a strong performance while introducing little extra computation cost.









\noindent \textbf{Person Selection Mechanism.}
In fact, the number of people usually varies in images. To this end, we introduce the person selection mechanism. As mentioned above, we set $N$ as the maximum number of person patches.  

During the training phase, we use the patch selection mechanism to facilitate the network training. For cases with persons less than $N$ in an input image, we simply calculate correlations between these persons or even a single person. But for parallel computing, we pad noise features to increase the person number to $N$  and mask them out to prevent self-attention calculation in Transformer blocks. When the number exceeds $N$, we randomly select a person as the target and fill the patches with the $N-1$ nearest neighbors.

In the testing stage, we apply the sliding window strategy for the inter-human relation module~(for the analysis of the maximum person number $N$, please see the ablation study). The sliding window divides all persons in the image into multiple $N$-person groups, and then we apply the inter-human relation module within each group.


\begin{table*}[h!]
\resizebox{.99\textwidth}{!}{
\begin{tabular}{cccccccccc}
\toprule
\multicolumn{1}{c|}{Method}                   & AP                       & AP$_{50}$                    & AP$_{75}$          & AR            & AR$_{50}$          & AR$_{75}$                     & AP$_{easy}$                 & AP$_{medium}$             & AP$_{hard}$              \\ 
\midrule
\multicolumn{10}{c}{Bottom-up}                                                                                                                                                                                                                                  \\ \hline
\multicolumn{1}{c|}{OpenPose~\cite{cao2019openpose}}                 & -                        & -                        & -             & -             & -             & -                        & 62.7                     & 58.7                     & 32.3                     \\
\multicolumn{1}{c|}{HigherHRNet~\cite{cheng2020higherhrnet}}              & 65.9                     & 86.4                     & 70.6          & -             & -             & -                        & 73.3                     & 66.5                     & 57.9                     \\
\multicolumn{1}{c|}{HigherHRNet Multi-scale~\cite{cheng2020higherhrnet}}  & 67.6                     & 87.4                     & 72.6          & -             & -             & -                        & 75.8                     & 68.1                     & 58.9                     \\
\multicolumn{1}{c|}{SPM~\cite{nie2019single}}                      & 63.7                     & 85.9                     & 68.7          & -             & -             & -                        & 70.3                     & 64.5                     & 55.7                     \\
\multicolumn{1}{c|}{DEKR~\cite{geng2021bottom}}                     & 65.7                     & 85.7                     & 70.4          & -             & -             & -                        & 73.0                       & 66.4                     & 57.5                     \\
\multicolumn{1}{c|}{DEKR Multi-scale~\cite{geng2021bottom}}         & 67.0                       & 85.4                     & 72.4          & -             & -             & -                        & 75.5                     & 68.0                       & 56.9                     \\
\multicolumn{1}{c|}{PINet~\cite{wang2021robust}}                    & 68.9                     & 88.7                     & 74.7          & -             & -             & -                        & 75.4                     & 69.6                     & 61.5                     \\
\multicolumn{1}{c|}{PINet Multi-scale~\cite{wang2021robust}}        & 69.9                     & 89.1                     & 75.6          & -             & -             & -                        & 76.4                     & 70.5                     & 62.2                     \\ \hline
\multicolumn{10}{c}{Top-down}                                                                                                                                                                                                                                   \\ \hline
\multicolumn{1}{c|}{Mask R-CNN~\cite{he2017mask}}               & 57.2                     & 83.5                     & 60.3          & 65.9          & 89.5          & 69.4 & 69.4 & 57.9 & 45.8 \\
\multicolumn{1}{c|}{AlphaPose~\cite{fang2017rmpe}}                & 61.0                       & 81.3                     & 66.0            & 67.6          & 86.7          & 71.8 & 71.2 & 61.4 & 51.1 \\
\multicolumn{1}{c|}{Simple baseline~\cite{xiao2018simple}}          & 60.8                     & 81.4                     & 65.7          & 67.3          & 86.3          & \multicolumn{1}{c}{71.8} & \multicolumn{1}{c}{71.4} & \multicolumn{1}{c}{61.2} & \multicolumn{1}{c}{51.2} \\
\multicolumn{1}{c|}{CrowdPose~\cite{li2019crowdpose}}                & 66.0                       & 84.2                     & 71.5          & 72.7          & 89.5          & \multicolumn{1}{c}{77.5} & \multicolumn{1}{c}{75.5} & \multicolumn{1}{c}{66.3} & \multicolumn{1}{c}{57.4} \\
\multicolumn{1}{c|}{OPEC-Net~\cite{qiu2020peeking}}                 & 70.6                     & 86.8                     & 75.6          & -             & -             & \multicolumn{1}{c}{-}    & \multicolumn{1}{c}{-}    & \multicolumn{1}{c}{-}    & \multicolumn{1}{c}{-}    \\
\multicolumn{1}{c|}{HRNet~\cite{sun2019deep}}                & 71.3                     & 91.1                     & 77.5          & -             & -             & \multicolumn{1}{c}{-}    & \multicolumn{1}{c}{80.5} & \multicolumn{1}{c}{71.4} & \multicolumn{1}{c}{62.5} \\
\multicolumn{1}{c|}{HRNet$\dag$~\cite{sun2019deep}}                & 72.8                     & 92.1                     & 78.7          & -             & -             & \multicolumn{1}{c}{-}    & \multicolumn{1}{c}{81.3} & \multicolumn{1}{c}{73.3} & \multicolumn{1}{c}{65.5} \\
\multicolumn{1}{c|}{TransPose-H~\cite{yang2021transpose}}            & 71.8                     & 91.5                     & 77.8          & 75.2          & 92.7          & \multicolumn{1}{c}{80.4} & \multicolumn{1}{c}{79.5} & \multicolumn{1}{c}{72.9} & \multicolumn{1}{c}{62.2} \\ 
\multicolumn{1}{c|}{HRFormer-B~\cite{YuanFHLZCW21}}            & 72.4                   & 91.5                    & 77.9        & 75.6          & 92.7          &  \multicolumn{1}{c}{81.0} & \multicolumn{1}{c}{80.0} & \multicolumn{1}{c}{73.5} & \multicolumn{1}{c}{62.4} \\ 
\hline

\multicolumn{1}{c|}{\textbf{$I^2$R-Net}~(Vanilla version, 1st stage: HRNet)}            & 72.3                     & 92.4                     & 77.9          & 76.5          & 93.2          & \multicolumn{1}{c}{81.9} & \multicolumn{1}{c}{79.9} & \multicolumn{1}{c}{73.2} & \multicolumn{1}{c}{62.8} \\ 
\multicolumn{1}{c|}{\textbf{$I^2$R-Net}~(1st stage: TransPose-H)} & 76.3 & 93.5 & 82.2 & 79.1 & 94.0 & \multicolumn{1}{c}{84.4} & \multicolumn{1}{c}{83.2} & \multicolumn{1}{c}{77.0} & \multicolumn{1}{c}{67.4} \\ 
\multicolumn{1}{c|}{\textbf{$I^2$R-Net}~(1st stage: HRFormer-B)}            & \textbf{77.4}                   & \textbf{93.6}                    & \textbf{83.3}        & \textbf{80.3}          & \textbf{94.5}          & \multicolumn{1}{c}{\textbf{85.5}} & \multicolumn{1}{c}{\textbf{83.8}} & \multicolumn{1}{c}{\textbf{78.1}} & \multicolumn{1}{c}{\textbf{69.3}} \\ 

\bottomrule
\end{tabular}
}
\caption{\label{tab:CrowdPose_Res}Comparisons with state-of-the-art methods on the CrowdPose dataset. The default input resolution for experiments is $256\times192$, while $\dag$ denotes the input resolution is $384\times288$.}
\end{table*}

\begin{table*}[htbp]
\centering
\Large
\resizebox{.999\textwidth}{!}{
\begin{tabular}{c|ccc|cccccccc}
\toprule
Method                      & Input size & \multicolumn{1}{c}{\#Params} & \multicolumn{1}{c|}{FLOPs}     & AP                 & AP$_{50}$          & AP$_{75}$          & AP$_{M}$           & AP$_{L}$           & AR           & AR$_{M}$           & AR$_{L}$ \\ 
\midrule
HRNet-W32~\cite{sun2019deep}       & 256×192    & 28.5M                        & 7.1G                          & 74.4               & 90.5          & 81.9          & 70.8          & 81.0          & 78.9     & 75.7  & 85.8 \\
HRNet-W48~\cite{sun2019deep}       & 256×192    & 63.6M                        & 14.6G                         & 75.1               & 90.6          & 82.2          & 71.5          & 81.8          & 80.4     & 76.2  & 86.4   
\\
\textbf{$I^2$R-Net}~(Vanilla version, 1st stage:HRNet-W48-S)        & 256×192    & 18.0M                        & 10.2G                         & 75.3               & 90.2          & 81.9          & 71.7          & 82.4          & 80.5     & 76.1 & 86.8     \\ \hline
TransPose-H~\cite{yang2021transpose}     & 256×192    & 17.5M                        & 21.8G                         & 75.8               & 90.1          & 82.1          & 71.9          & 82.8          & 80.8     & 76.4  & 87.2     \\
\textbf{$I^2$R-Net}~(1st stage:TransPose-H)        & 256×192    & 18.0(+0.5)M          & 22.7(+0.9)G                   & 75.8      & 90.4     & 82.1     & 72.0      & 82.9      & 80.9      & 76.6  & 87.3      \\ \hline
HRFormer-B$\star$~\cite{YuanFHLZCW21}       & 256×192    & 43.2M                       & 12.2G                         & 75.6               & 90.8          & 82.8          & 71.7          & 82.6          & 80.8      & 76.5  & 87.2    \\
\textbf{$I^2$R-Net}~(1st stage:HRFormer-B$\star$)        & 256×192    & 43.7(+0.5)M                  & 12.8(+0.6)G                   & 76.4               & 90.8          & 83.2          & 72.3          & 83.7          & 81.4     & 76.9 & 88.1     \\
\hline
HRFormer-B$\star$~\cite{YuanFHLZCW21}  & 384×288    & 43.2M                       & 26.8G                         & 77.1               & 90.9          & 83.5          & 69.8          & 80.5          & 82.0      & 77.7  & 88.3 \\ 
\textbf{$I^2$R-Net}~(1st stage:HRFormer-B$\star$)       & 384×288    & 43.7(+0.5)M                 &  29.6(+2.8)G                  & \textbf{77.3}      &\textbf{91.0}  &\textbf{83.6}  &\textbf{73.0}  & \textbf{84.5} & \textbf{82.1}  & \textbf{77.7} & \textbf{88.6} \\ 
\bottomrule
\end{tabular}
}
\caption{\label{tab:COCO_Res}Results on the COCO pose estimation \emph{validation} set. The results of methods marked with $\star$ are obtained by running the official code and model. Following TransPose, HRNet-W48-S only uses early 3 stages of HRNet-W48.}
\end{table*}

\section{Experiments}



\subsection{Datasets and Evaluation Metrics}

\paragraph{CrowdPose}~\cite{li2019crowdpose} is one of the most commonly used datasets in multi-person pose estimation task. The images in this dataset are heavily crowded and challenging. It contains 10K, 2K and 8K images for training set, validation set and testing set, respectively. Following \cite{cheng2020higherhrnet,geng2021bottom}, we train our models on the training set and validation set, and evaluate on the testing set.

\paragraph{OCHuman}~\cite{zhang2019pose2seg} is a recently proposed benchmark to examine the limitations of MPPE in highly occluded scenarios. It consists of 4731 images, including 2500 images for validation and 2231 images for testing. Following \cite{qiu2020peeking}, we train models on the validation set and report the performance on the testing set.

\paragraph{COCO}~\cite{lin2014microsoft} is one popular benchmark for keypoint detection. Although it contains few crowded scenes, we use it to validate the generation ability of our method for non-crowded scenes. We train the models on the training set and report the results on the validation set.


\paragraph{Evaluation Metrics.} We follow the standard evaluation procedure and use OKS-based metrics for pose estimation~\cite{lin2014microsoft}. We report average precision~(AP), average recall~(AR) with different thresholds. Following \cite{li2019crowdpose}, on CrowdPose, we also report AP$_{easy}$, AP$_{medium}$ and AP$_{hard}$, which refers to the performance under different crowd index.

\subsection{Implementation Details}
We use Adam optimizer and the cosine annealing learning rate decay from 1e-4 to 1e-5. We also adopt the coordinate decoding strategy in \cite{zhang2020distribution} to reduce the quantisation error when decoding from downscaled heatmaps. For the training batch size, we choose 64 for $I^2$R-Net at  $256 \times 192$ resolution, and 32 at $384 \times 288$ resolution. The length of each token channel $d$ is 96. The number of transformer layers $M=6$ in the vanilla $I^2$R-Net, and $M=2$ for other models. Each experiment takes $8\times$ 32G-V100 GPUs for training.


\begin{table}[htp]
\centering
\Large
\resizebox{.48\textwidth}{!}{
\begin{tabular}{cccc}
\toprule
\multicolumn{1}{c|}{Method}                & AP            & AP$_{50}$          & AP$_{75}$          \\ 
\midrule
\multicolumn{4}{c}{Bottom-up}                                                              \\ \hline
\multicolumn{1}{c|}{SPM~\cite{nie2019single}}                       & 47.6          & 67.5          & 53.2          \\
\multicolumn{1}{c|}{HigherHRNet~\cite{cheng2020higherhrnet}}        & 27.7          & 66.9          & 15.9          \\
\multicolumn{1}{c|}{DEKR~\cite{geng2021bottom}}                     & 52.2          & 69.9          & 56.6          \\
\multicolumn{1}{c|}{PINet~\cite{wang2021robust}}                    & 59.8          & 74.9          & 65.9          \\ \hline
\multicolumn{4}{c}{Top-down}                                                               \\ \hline
\multicolumn{1}{c|}{Mask R-CNN~\cite{he2017mask}}                   & 20.2          & 33.2          & 24.5          \\
\multicolumn{1}{c|}{SimplePose~\cite{xiao2018simple}}               & 24.1          & 37.4          & 26.8          \\
\multicolumn{1}{c|}{CrowdPose~\cite{li2019crowdpose}}               & 27.5          & 40.8          & 29.9          \\
\multicolumn{1}{c|}{OPEC-Net~\cite{qiu2020peeking}}                 & 29.1          & 41.3          & 31.4          \\
\multicolumn{1}{c|}{HRFormer-B~\cite{YuanFHLZCW21}}                 & 62.1          & 81.4          & 67.1          \\
\multicolumn{1}{c|}{ThansPose-H~\cite{yang2021transpose}}         & 62.3          & 82.7          & 67.1          \\ \hline
\multicolumn{1}{c|}{\textbf{$I^2$R-Net}~(Vanilla version, 1st stage: HRNet)}        & 64.3          & 85.0          & 69.2    \\
\multicolumn{1}{c|}{\textbf{$I^2$R-Net}~(1st stage: HRFormer-B)}   & 66.5          & 83.8          & 71.4         \\
\multicolumn{1}{c|}{\textbf{$I^2$R-Net}~(1st stage:TransPose-H)} & \textbf{67.8} & \textbf{85.0} & \textbf{72.8} \\ 
\bottomrule
\end{tabular}
}
\caption{\label{tab:OCHuman_Res}Comparisons with state-of-the-art methods on the OCHuman \emph{testing} set after training on OCHuman \emph{validation} set.}

\end{table}

\begin{figure*}[ht!]
 \centering
    \includegraphics[width=0.98\textwidth]{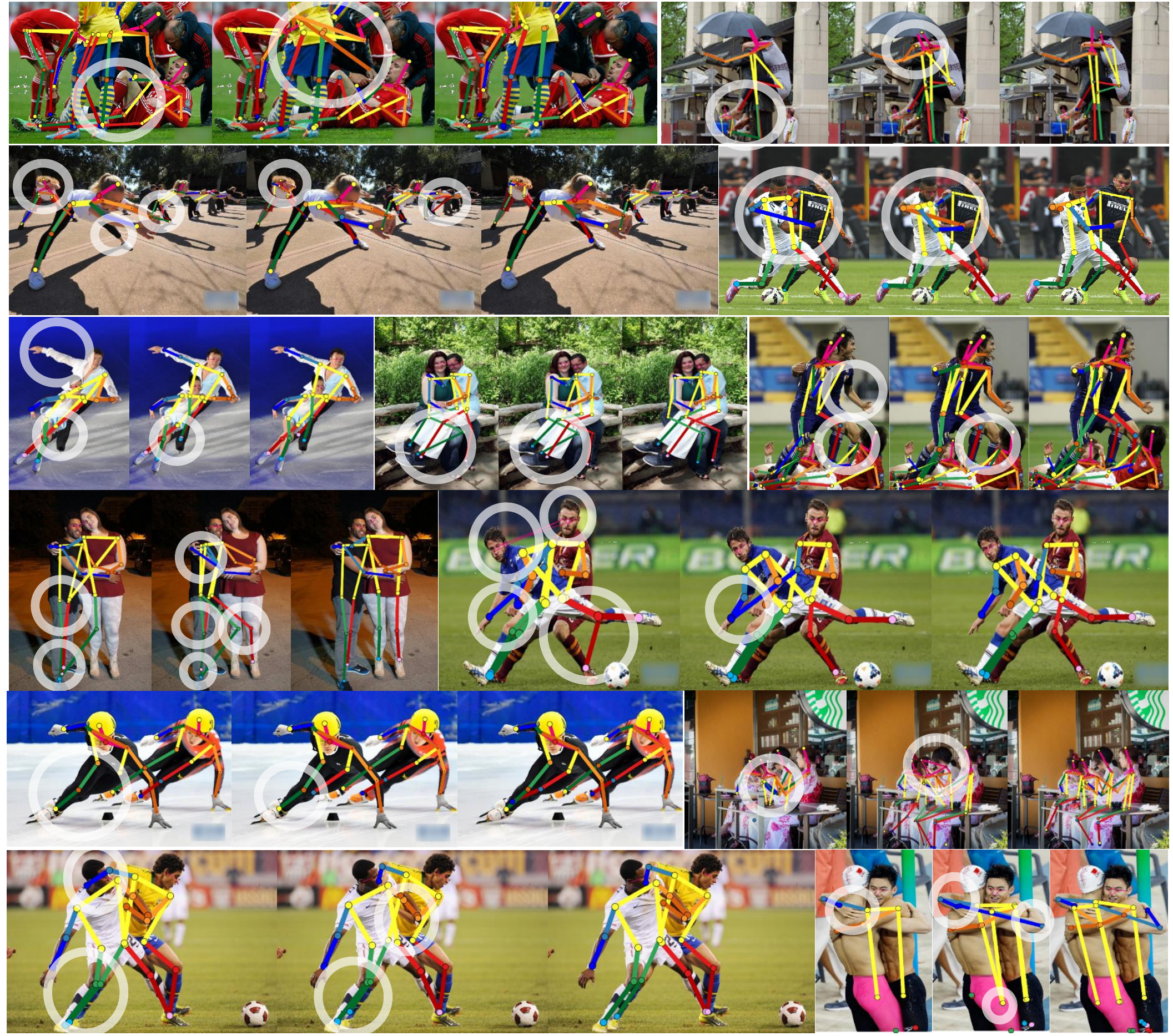}
  \caption{\textbf{Visualization Comparisons.} From left to right, each example shows the results of TransPose, HRFormer, and $I^2$R-Net. The incorrect regions are marked with white circles.\label{vis_res}}
\end{figure*}

\begin{figure}[t]
 \centering
    \includegraphics[width=0.48\textwidth]{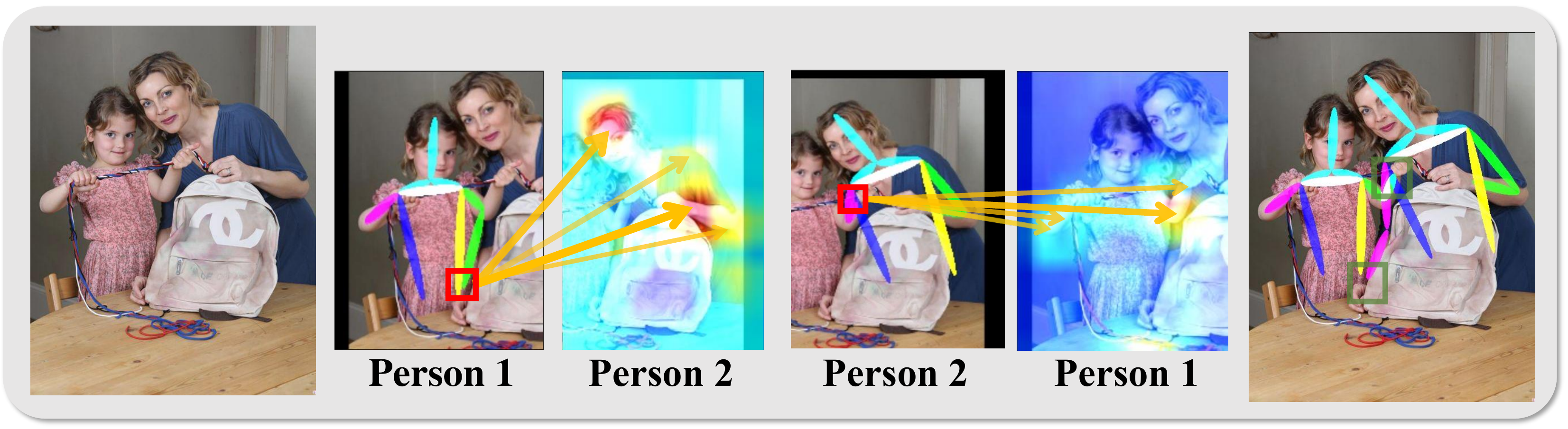}
  \caption{\textbf{Inter-Human Attention Visualization.} The incorrect region~(marked in the red box) is corrected with the aid of higher response regions, which are indicated by the orange arrows.\label{att_vis}}
\end{figure}

\subsection{Compare with the State-of-the-art Methods}


\paragraph{Results on the CrowPose testing set.}
As shown in Table~\ref{tab:CrowdPose_Res}, thanks to our Inter-human relation module, the $I^2$R-Net~(1st stage: TransPose-H) achieves 76.3\% AP, which outperforms TransPose-H. Moreover, the $I^2$R-Net~(1st stage: HRFormer-B) achieves 77.4\% AP at $256 \times 192$ resolution, even surpassing the performance of HRNet$\dag$ at resolution of $384 \times 288$. This demonstrates that our method is very suitable for crowded scenes, and performs better than other methods even without bells and whistles.

\paragraph{Results on the OCHuman testing set.}
On OCHuman dataset, our method also obtains superior performance. The results are shown in Table~\ref{tab:OCHuman_Res}. The vanilla $I^2$R-Net achieves 64.3\% AP on OCHuman test set, which is comparable with state-of-the-art methods. Furthermore, it is clear in Table~\ref{tab:OCHuman_Res} that, equipped with the Inter-Human Relation Module, HRFormer-B and TransPose-H are significantly improved, reaching 66.5\% and 67.8\% AP, respectively. 
    
\paragraph{Results on the COCO pose estimation validation set.} As shown in Table~\ref{tab:COCO_Res}, $I^2$R-Net achieves 77.3\% AP, which is competitive against state-of-the-art methods. The performance gain on COCO is not that large compared with OCHuman and CrowdPose. This is because that COCO contains around 48\% single-person images. Hence, our method for multi-person scenarios is not able to give full play. But this result demonstrates that our method also works well in non-crowded scenes.




\subsection{Ablation Studies}

\paragraph{Effectiveness of Inter-Human Relation Module.} 
On the CrowdPose and OCHuman, $I^2$R-Net outperforms all existing methods at the same resolution. On COCO datasets, $I^2$R-Net also achieves competitive performance against state-of-the-art methods, especially outperforming on AP$_{M}$ and AP$_{L}$. As shown in Table~\ref{tab:CrowdPose_Res}, under our two-stage framework, TransPose can significantly surpass HRNet$\dag$ with Inter-Human Relation Module even at lower resolution. This demonstrates that our Inter-Human Relation Module can bring significant performance improvements.

As is discussed in Section~\ref{InterModule}, the relation modeling capability of inter-human relation module is not limited to cross-instance, but can also relates parts in-instance. This can be evidenced in Table \ref{tab:CrowdPose_Res}, \ref{tab:COCO_Res} and \ref{tab:OCHuman_Res}, where the vanilla Inter-Human Relation~(without intra- stage, and extract per-person feature using a backbone) achieves comparable performance with state-of-the-art methods.


\paragraph{Instance patch number.}
An important factor of $I^2$R-Net is the number of instances fed into the Inter-human relation module. We compare the results of different instance numbers on CrowdPose in Table~\ref{tab:Ablation_patch_number}. We can observe that the AP value tends to be saturated when $N$ is 4 and reaches the highest when $N$ is 6. A higher or lower number of patches will cause a drop in AP value. This is because too few patches may lose information between persons, while too many patches introduce unnecessary cross-human correlations.

\paragraph{Effectiveness of Intermediate-supervision.} In our framework, we use MSE loss in the first stage as intermediate-supervision~(denoted as IS). As shown in the second and last rows in Table~\ref{tab:Ablation_pe_intermediate}, the intermediate-supervision improves the performance.

\paragraph{Position encoding.}\label{PE_ablation}
We also try to add position encoding into $I^2$R-Net. As shown in Table~\ref{tab:Ablation_pe_intermediate}, neither using sinusoidal mode nor encoding the global position as a positional encoding can improve the results. One possible reason is that the Intra-Human Relation Module has learned the local correlations between parts of each person, and the global interactions between instances are position-independent.

\paragraph{Person selection mechanism.}\label{PatchExtractionStrategy} We compare two strategies for person selection. One is the nearest neighbor strategy, which randomly selects a target person and selects $N-1$ closest neighbors; the other is to completely randomly select $N$ persons. As shown in Table~\ref{tab:Ablation_patch_strategy}, there are no significantly difference between them, which means the multi-person interaction is position-free. 

\paragraph{Token size setting.} In Table~\ref{tab:label subtable D}, the setting of $16 \times 12 \times N$ achieves a better result. It shows that our method can learn enough inter-human relations at low resolution, but $1 \times 1 \times N$ leads to low precision due to the lack of spatial information.

\begin{table}[h]
    \small
	\begin{subtable}[h]{0.22\textwidth}
		\centering
        \begin{tabular}{c|cccc}
        \toprule
        Patch & AP            & AR               \\ 
        \midrule
        1     & 74.6          & 77.5             \\
        2     & 75.7          & 78.7             \\
        4     & 76.1          & 79.0             \\
        6     & \textbf{76.3} & \textbf{79.1}    \\
        8     & 76.2          & 79.1             \\ 
        \bottomrule
        \end{tabular}
		\caption{Instance patch number.}
		\label{tab:Ablation_patch_number}
	\end{subtable}
	\hfill
	\begin{subtable}[h]{0.22\textwidth}
		\centering
		\begin{tabular}{c|c}
        \toprule
        Model          & AP        \\ 
        \midrule
        Base           & 66.2      \\
        Base+IS        & \textbf{66.5}      \\
        Base+pe$_{sine}$        & 66.2      \\
        Base+pe$_{global}$        & 65.3      \\
        Base+pe$_{global}$+IS     & 65.7      \\ 
        \bottomrule
        \end{tabular}
		\caption{Effectiveness of components in $I^2$R-Net.}
		\label{tab:Ablation_pe_intermediate}
	\end{subtable}
	\hfill
	\begin{subtable}[h]{0.22\textwidth}
		\centering
		\begin{tabular}{c|cc}
        \toprule
        Strategy & AP & AR     \\ 
        \midrule
        Nearest        & 76.0 & 79.0 \\
        Random         & 76.0 & 78.9  \\ 
        \bottomrule
        \end{tabular}
		\caption{Person selection strategy.}
		\label{tab:Ablation_patch_strategy}
	\end{subtable}
	\hfill
	\begin{subtable}[h]{0.22\textwidth}
		\centering
		\begin{tabular}{c|c}
        \toprule
        Token Size & AP   \\ 
        \midrule
        $16 \times 12 \times N$      & \textbf{66.5} \\
        $32 \times 24 \times N$       & 66.4 \\
        $1 \times 1 \times N$           & 66.0 \\ 
        \bottomrule
        \end{tabular}
		\caption{Token size setting.}
		\label{tab:label subtable D}
	\end{subtable}
	\caption{Ablation study of our $I^2$R-Net with different settings.}
	\label{tab:Ablation_Study}
\end{table}


\subsection{Qualitative Analysis}
In Figure~\ref{vis_res}, we visualize our results compared with two competitive state-of-the-art methods ~(TransPose and HRFormer). The proposed method is able to handle the heavily crowded cases and localize the key points correctly. Even though in occluded scenes, we can also infer plausible poses.

By visualizing the attention maps between persons, the performance gain can be justified. As shown in Figure~\ref{att_vis}, the incorrect localized keypoints are corrected by highly related regions of the other person.


\subsection{Model Size and Computation Cost}
As shown in Table~\ref{tab:COCO_Res}, Inter-Human Relation Module is lightweight yet effective. In the two-stage framework, for instance, HRFormer with Inter-Human Relation Module merely increased the parameters by 0.5M and FLOPS by 0.6G. This demonstrates that our Inter-Human Relation Module cost a few and acceptable extra computations but achieves significant improvement for multi-person estimation. 

\begin{figure}
    \centering
    \includegraphics[width=0.48\textwidth]{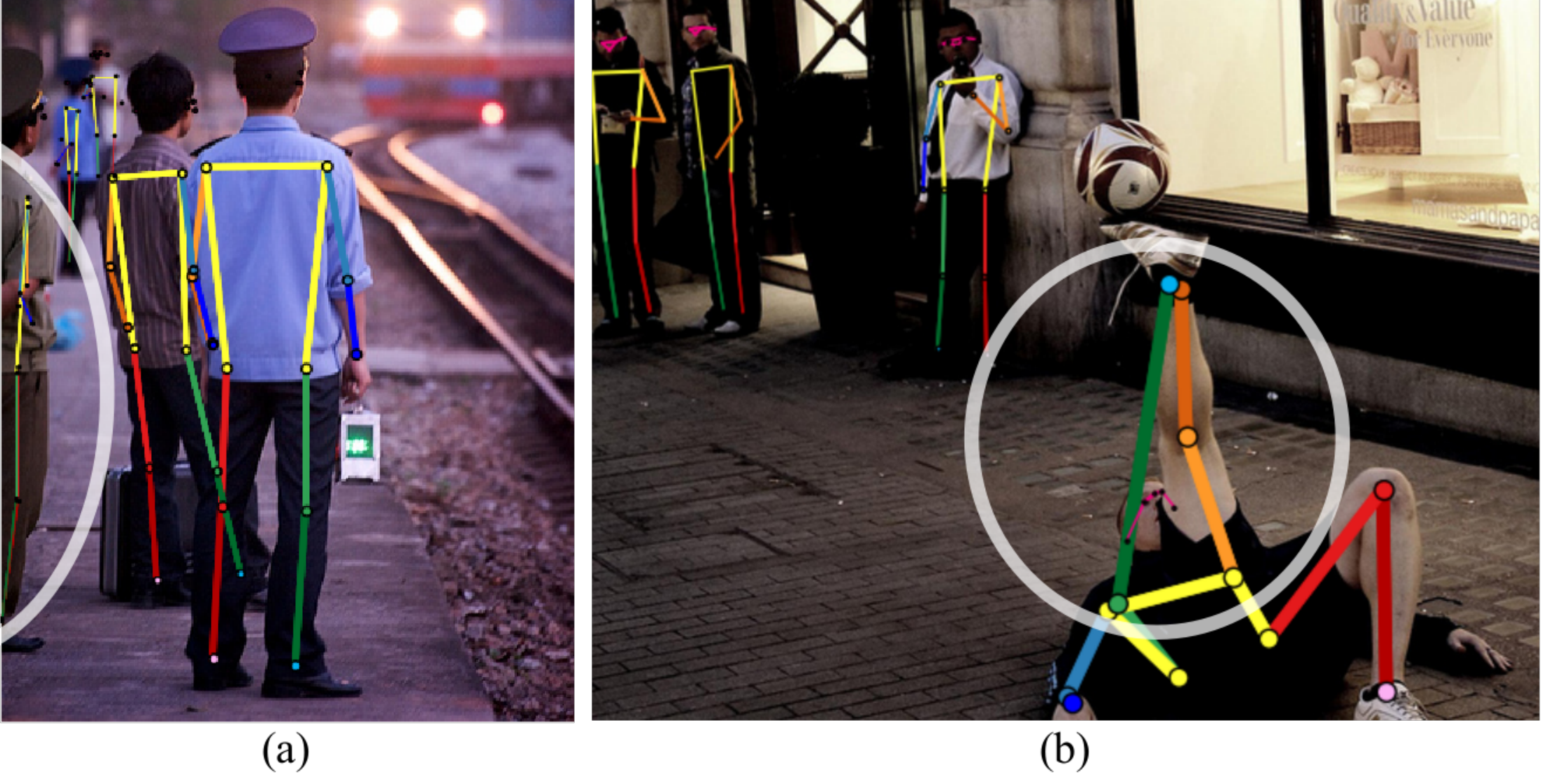}
    \caption{\textbf{Failure cases.} The incorrect regions are marked with white circles.}
    \label{fig:failure}
    \vspace{-2mm}
\end{figure}

\subsection{Failure Cases}
Although our approach outperforms the state-of-the-art methods, we still suffer from several extremely challenging cases. Like previous work, our method may fail in cases where only part of the person is visible (Figure~\ref{fig:failure}(a)) or the inter-human relationships are insignificantly correlated (Figure~\ref{fig:failure}(b)). All these cases with poor relations may lead to failure predictions.

\section{Concluding Remarks}


This work investigates and evidences that collaboratively estimating poses via modeling multi-person correlation promotes pose estimation. We propose a flexible and efficient framework to integrate intra-human and inter-human information, leading to superior results than state-of-the-art methods. Inspired by the idea of this work, promising future work is to extend the inter-human relation from the spatial dimension to the temporal dimension. Another interesting direction is to explore the performance gain via correlating the human-object relationships in various vision tasks.

\section*{Acknowledgements}
This work was supported in part by NSFC~(No. 62072382) and Yango Charitable Foundation.

\appendix


\bibliographystyle{named}
\bibliography{ijcai22}

\end{document}